\documentclass[letterpaper, 10 pt, conference]{ieeeconf}  

\IEEEoverridecommandlockouts                              

\usepackage{graphics} 
\usepackage{epsfig} 
\usepackage{amsmath} 
\usepackage{amssymb}  
\usepackage{bm}
\usepackage{flushend}
\usepackage{soul}
\newcommand{\argmax}{\mathop{\mathrm{argmax}}}

\newcommand{\maxover}{\mathop{\mathrm{max}}} 

\title{\LARGE \bf
DICE: Diverse Diffusion Model with Scoring for Trajectory Prediction
}

\author{Younwoo Choi$^{1}$,  Ray Coden Mercurius$^{2}$, Soheil Mohamad Alizadeh Shabestary$^{2}$, Amir Rasouli$^{2}$
\thanks{$^{1}$ University of Toronto. Work done while at Huawei. {\tt\small ywchoi@cs.toronto.edu}}
\thanks{$^{2}$Noah's Ark Laboratory, Huawei, Canada. {\tt\small first.last@huawei.com}%
}}
\begin{document}
\maketitle
\thispagestyle{empty}
\pagestyle{empty}

\begin{abstract}
Road user trajectory prediction in dynamic environments is a challenging but crucial task for various applications, such as autonomous driving. One of the main challenges in this domain is the multimodal nature of future trajectories stemming from the unknown yet diverse intentions of the agents. Diffusion models have shown to be very effective in capturing such stochasticity in prediction tasks. However, these models involve many computationally expensive denoising steps and sampling operations that make them a less desirable option for real-time safety-critical applications. To this end, we present a novel framework that leverages diffusion models for predicting future trajectories in a computationally efficient manner. To minimize the computational bottlenecks in iterative sampling, we employ an efficient sampling mechanism that allows us to maximize the number of sampled trajectories for improved accuracy while maintaining inference time in real time. Moreover, we propose a scoring mechanism to select the most plausible trajectories by assigning relative ranks. We show the effectiveness of our approach by conducting empirical evaluations on common pedestrian (UCY/ETH) and autonomous driving (nuScenes) benchmark datasets on which our model achieves state-of-the-art performance on several subsets and metrics.
\end{abstract}

\section{Introduction}
Accurate prediction of road user behavior is prerequisite to safe motion planning in autonomous driving systems. One of the key challenges in trajectory prediction is the probabilistic and multimodal nature of road users’ behaviors. To model such uncertainty, many approaches have been proposed, such as Generative Adversarial Networks (GANs) \cite{dendorfer2021mggan,Fang_2020_CVPRtpnet,gupta2018socialsocialgan,sun2020reciprocal}, Conditional Variational Autoencoder (CVAE) \cite{chen2022personalized,ivanovic2019trajectron,lee2017desire,Liu_2021_ICCV,salzmann2021trajectron}, anchor-based proposal networks \cite{varadarajan2021multipath}, or target (intention) prediction networks \cite{choi2020drogon,zhao2020tnt,gu2021densetnt}. However, these methods are not without challenges, encompassing issues, such as unstable training, artificial dynamics within predicted trajectories, and reliance on hand-crafted heuristics that lack generalizability. 

Diffusion models have recently gained popularity as powerful tools for various generative tasks in machine learning  \cite{rombach2022high, saharia2022photorealistic, ho2022imagen, yang2022diffusion}. These models learn the process of transforming informative data into  Gaussian noise and how to reversely generate meaningful output from noisy data. Although effective, diffusion models impose high computational costs due to successive denoising and sampling operations making their performance limited for real-time applications, such as trajectory prediction. 

To this end, we propose a novel computationally efficient diffusion-based model for road user trajectory prediction. Our approach benefits from the efficient sampling method, Denoising Diffusion Implicit Models (DDIM) \cite{song2022denoisingddim} resulting in $20\times$ computational speed-up, allowing us to oversample from trajectory distributions in order to maximize the diversity and coverage of predicted trajectories, while maintaining the inference speed well below existing approaches. To select the most likely trajectory candidates, we propose a novel scoring network that assigns relative rankings in conjunction with a non-maximum suppression operation in order to downsample the trajectories into the final prediction set. To highlight the effectiveness of our approach, we conduct extensive empirical studies on common trajectory prediction benchmark datasets, UCY/ETH \cite{ucy,4409092eth} and nuScenes \cite{nuscenes}, and show our model achieves state-of-the-art performance on some subsets and metrics while maintaining real-time inference time. 

\section{Related Work}
\subsection{Trajectory Prediction}
Trajectory prediction is modeled as a sequence prediction problem where the future of the agents is predicted based on their observed history and potentially available contextual information. In the pedestrian prediction domain, one of the key challenges is to model the interactions among pedestrians for better estimates of their future behavior. These methods include, spatial pooling of representations \cite{alahi2016social}, graph architectures \cite{Huang_2019_ICCV,ivanovic2019trajectron,mohamed2020socialstgcnn,salzmann2021trajectron,sun2020recursive,yu2020spatiotemporal}, attention mechanisms \cite{fernando2017soft,kosaraju2019socialbigat,sadeghian2018sophie,vemula2018social,zhang2019srlstm}, and in egocentric setting, semantic scene reasoning \cite{Rasouli_2021_ICCV,rasouli2023pedformer}.  In the context of autonomous driving, it is also important to model agent-to-map interactions. For this purpose, models rely on environment representations in the form of drivable areas \cite{park2020diverse, amirloo2022latentformer}, rasterized maps \cite{salzmann2021trajectron,gilles2021gohome}, point-clouds \cite{ye2021tpcn}, and computationally efficient vectorized representations in conjunction with graph neural networks \cite{gao2020vectornet,gu2021densetnt}, or transformers \cite{hivt, nayakanti2023wayformer} for generating holistic representations of the scenes and interactions.

Another main challenge in trajectory prediction is to capture the uncertainty and multi-modality in the agent's behaviour. To address this problem, a category of models resort to explicitly predicting the goal (intentions or target) of agents and predict future trajectories conditional on those goals, which are typically defined using heuristic methods, which limit the generalizability of these approaches \cite{zhao2020tnt,gu2021densetnt}. Other methods, such as Generative Adversarial Networks (GANs) \cite{dendorfer2021mggan,Fang_2020_CVPRtpnet,gupta2018socialsocialgan,sun2020reciprocal} and Conditional Variational Autoencoders (CVAEs) \cite{chen2022personalized,ivanovic2019trajectron,lee2017desire,Liu_2021_ICCV,salzmann2021trajectron} implicitly capture agents' intentions. These methods introduce latent variables that are randomly sampled from a simple distribution to produce complex and multi-modal distributions for the predicted trajectories. However, existing generative models suffer from limitations, such as mode collapse, unstable training, or the generation of unrealistic trajectories \cite{thanhtung2020catastrophic,zhao2017deeper}, highlighting the need for more robust and accurate models. 

\subsection{Denoising Diffusion Models}
Denoising Diffusion Probabilistic Models (DDPM) \cite{ho2020denoisingddpm}, commonly referred to as diffusion models have gained popularity various generative tasks such as image \cite{rombach2022high, saharia2022photorealistic}, audio \cite{kong2021diffwave}, video \cite{ho2022imagen, yang2022diffusion}, and 3D point cloud generation \cite{luo2021diffusion3dcloud}. These models simulate a diffusion process motivated by non-equilibrium thermodynamics, where a parameterized Markov chain is learned to gradually transition from a noisy initial state to a specific data distribution. More recently, diffusion methods have been adopted in trajectory generation and prediction tasks \cite{gu2022stochasticmid, jiang2023motiondiffuser, rempe2023trace}. The approach in \cite{gu2022stochasticmid} models the indeterminacy of human behaviour using a transformer-based trajectory denoising module to capture complex temporal dependencies across trajectories. The authors of \cite{rempe2023trace} introduce a model for generating realistic pedestrian trajectories that can be controlled to meet user-defined goals by implementing a guided diffusion model. MotionDiffuser introduced in \cite{jiang2023motiondiffuser} creates a diffusion framework for multi-agent joint prediction with optional attractor and repeller guidance functions to enforce compliance with prior knowledge, such as agent intention and accident avoidance.

Although diffusion-based models have proven effective, there exist critical shortcomings to their adoption in practice. For instance, the inference time of these models is highly computationally expensive due to an iterative denoising algorithm that requires a large number of forward passes. For example, MotionDiffuser's inference latency is 408.5ms (32 diffusion steps) compared to the conventional prediction models, such as HiVT's \cite{hivt} with 69ms latency. This characteristic makes diffusion models a less desirable option for real-time safety-critical applications, such as autonomous driving. To speed-up prediction inference time, Leapfrog Diffusion Model (LED) \cite{mao2023leapfrog} is proposed that learns to skip a large number of denoising steps in order to accelerate inference speed. However, LED is only effective when dealing with small dimensional trajectory data without any complex contextual encoding. To address this shortcoming, in the proposed approach we focus on improving efficiency at the sampling stage providing a more general framework for different data representation.

\noindent\textbf{Contributions} of this paper are threefold: 1) We propose a novel diffusion-based model for trajectory prediction that relies on efficient sampling operation for over-sampling trajectories and a novel scoring mechanism for relatively ranking them to produce final prediction set; 2) we conduct empirical evaluation by comparing the proposed approach to the past arts and highlight the effectiveness of our approach in both pedestrian and autonomous driving prediction domains; 3) We conduct ablation studies on the effect of proposed scoring scheme and oversampling on the accuracy of prediction and inference time.


   \begin{figure*}[thpb]
      \centering
      \vspace{+0.2cm}
      \includegraphics[scale=1]{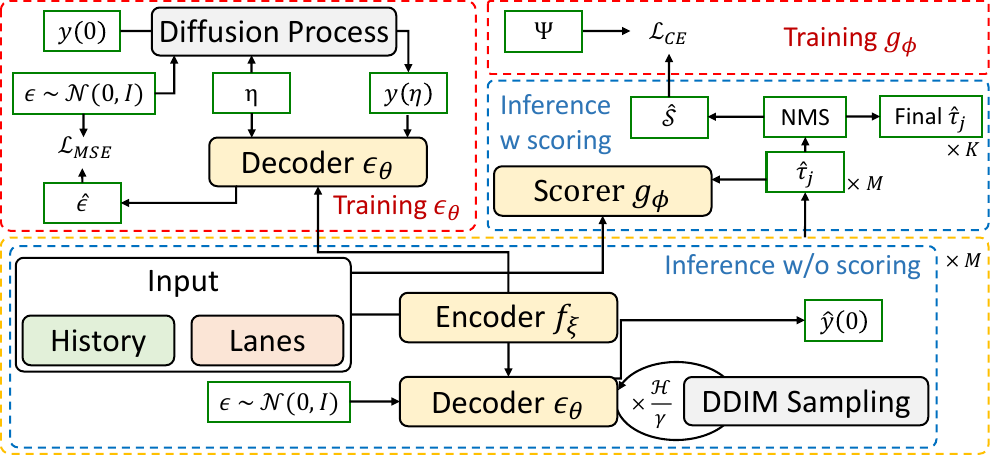}
      \caption{Overview of the proposed framework. During training of decoder $\bm{\epsilon}_{\theta}$, encoder $f_{\xi}$ encodes the history and the map into a feature embedding. Using diffusion step $\eta$, the feature embedding, and noisy $\mathbf{y}_{(\eta)}$, decoder $\bm{\epsilon}_{\theta}$ generates noise $\epsilon$ to noise clean $\mathbf{y}_{(0)}$ into $\mathbf{y}_{(\eta)}$. In the second stage, scoring network $g_{\phi}$ takes $M$ generated trajectories $\hat{\bm{\tau}}_j$ using efficient DDIM sampling method along with the feature embedding from the encoder. A non-maximum suppression algorithm is applied to trajectories sorted in descending order based on their predicted scores, from which final $K$ trajectories are selected.} 
      \label{fig:architecture}\vspace{-0.4cm}
   \end{figure*}

\section{Methodology}
\subsection{Problem Formulation}
We represent the future trajectory of an agent $i$, $\bm{\tau}^{i}_{future}=[\bm{p}_{t+1}^{i},\ldots,\bm{p}_{t+T_f}^{i}]$ over $T_f$ time steps where $\bm{p}^i\in\mathbb{R}^2$ is the 2D coordinates of the agent. Similarly, the agent's trajectory over the last $T_p$ time steps is $\bm{\tau}^{i}_{past}=[\bm{p}_{t-T_p+1}^{i},\ldots,\bm{p}_{t}^{i}]$. Here, the objective is to learn distribution $p(\tau_{future}^i|\tau_{past}^i)$. 
\subsection{Architecture}
An overview of the proposed framework is illustrated in Figure \ref{fig:architecture}. Our model consists of an input layer containing trajectory history and lane information (if map information is available); An encoder inspired by \cite{hivt} that encoded interactions between agents and agents and road lanes (if map is available); A decoder based on \cite{gu2022stochasticmid} comprised of multiple transformer layers. The decoder is trained to generate meaningful trajectories from noisy data conditioned on the scene context and used iteratively in each step of the denoising; And lastly, an attention-based scorer ranks the generated trajectories. In the following subsections, we describe the key components of the proposed model.

\subsection{Data Processing}
Inspired by \cite{hivt}, we use translation- and rotation-invariant scene representation vectors by converting absolute positions to relative positions and rotating them according to the heading angle of each agent denoted by $\theta^i$ at the current timestamp $t$. We convert the trajectory coordinates into displacements, and rotate them so that all agents have the same heading at time step zero. Each trajectory is represented as $\mathbf{x}^i=[0, \mathbf{R}_i^T(\bm{p}_{t-T_p+2}^{i}-\bm{p}_{t-T_p+1}^{i}), \ldots, \mathbf{R}_i^T(\bm{p}_{t}^{i}-\bm{p}_{t-1}^{i})]$, where $\mathbf{R}_i\in\mathbb{R}^{2\times2}$ is the rotation matrix parameterized by $\theta^i$. $\mathbf{X} = [\mathbf{x}^1,\ldots,\mathbf{x}^N]\in\mathbb{R}^{N\times T_p\times2}$ is the set of agents' observations, where $N$ denotes the number of agents in a scene. Using this representation makes the model robust to both translation and rotation and consequently requires less training data as no rotation-augmented data is needed as in the case of other methods. Contrary to the standard approach in the literature, we also convert future trajectories $\bm{\tau}^{i}_{future}$ to the displacement of relative positions rotated and centered at the last point of the history, that is $\mathbf{y}^i=[\mathbf{R}_i^T(\bm{p}_{t+1}^{i}-\bm{p}_{t}^{i}),\ldots,\mathbf{R}_i^T(\bm{p}_{t+T_f}^{i}-\bm{p}_{t+T_f-1}^{i})]$.

\subsection{Conditional Diffusion Model for Trajectory Prediction}
The forward diffusion process is a Markov chain that gradually adds Gaussian noise to a sample drawn from the data distribution, $\mathbf{y}_{(0)}\sim q(\mathbf{y})$, iteratively for $H$ times, in order to produce progressively noisier samples. The approximate posterior distribution is given by,

$$
q(\mathbf{y}_{(1:H)}|\mathbf{y}_{(0)}) = \prod^H_{\eta=1}q(\mathbf{y}_{(\eta)}|\mathbf{y}_{(\eta-1)}) \eqno{(1)}
$$ 
$$
q(\mathbf{y}_{(\eta)}|\mathbf{y}_{(\eta-1)}) = \mathcal{N}(\mathbf{y}_{(\eta)};\sqrt{1-\beta_{\eta}}\mathbf{y}_{(\eta-1)},\beta_{\eta}\mathbf{I}) \eqno{(2)}
$$
\noindent where $H$ denotes the total number of diffusion steps and $\beta_{\eta}$ is a uniformly increasing variance to control the level of noise. By setting $\alpha_{\eta}=1-\beta_{\eta}$ and $\bar{a}_{\eta}=\prod^{\eta}_{s=1}\alpha_s$, we get,
$$
q(\mathbf{y}_{(\eta)}|\mathbf{y}_{(0)}) = \mathcal{N}(\mathbf{y}_{(\eta)};\sqrt{\bar{\alpha}}_{\eta}\mathbf{y}_{(0)},(1-\bar{\alpha}_{\eta})\mathbf{I}) \eqno{(3)}.
$$
Here, if $H$ is large enough, $q(\mathbf{y}_{(H)})\sim\mathcal{N}(\mathbf{y}_{(H)};\bm{0},\mathbf{I})$, where $\mathcal{N}$ is a normal distribution.

For future trajectory prediction, we apply the reverse diffusion process by reducing noise from trajectories sampled under the noise distribution. The initial noisy trajectory is sampled from a normal distribution $\mathcal{N}(\mathbf{y}_{(H)};\bm{0},\mathbf{I})$, and then  it is iteratively run through conditioned denoising transition $p_{\theta}(\mathbf{y}_{(\eta-1)}|\mathbf{y}_{(\eta)}, \bm{C})$ parameterized by $\theta$ for $H$ steps. Here, $\bm{C}\in\mathbb{R}^{d_c}$ denotes a feature embedding with the dimension of $d_c$.  representing the scene context, learned by encoder $f_{\xi}$ parameterized by $\xi$. Formally, the process is as follows,
$$
p_{\theta}(\mathbf{y}_{(0:H)}) = p(\mathbf{y}_{(H)})\prod^H_{\eta=1}p_{\theta}(\mathbf{y}_{(\eta-1)}|\mathbf{y}_{(\eta)}) \eqno{(4)}
$$
At each step $\eta$, we have
$$
p_{\theta}(\mathbf{y}_{(\eta-1)}|\mathbf{y}_{(\eta)},\bm{C}) = \mathcal{N}(\mathbf{y}_{(\eta-1)};\bm{\mu}_{\theta}(\mathbf{y}_{(\eta)},\eta,\bm{C}),\bm{\Sigma}_{(\eta)}) \eqno{(5)}
$$where $p(\mathbf{y}_{(H)}) = \mathcal{N}(\mathbf{y}_{(H)};\bm{0},\mathbf{I})$ and $\bm{\Sigma}_{(\eta)}$ is a fixed variance scheduler, $\beta_{(\eta)}\mathbf{I}$.


\subsection{Scoring Network}
Generative models, such as CVAEs \cite{Sohn2015LearningSOcvae} and diffusion models \cite{ho2020denoisingddpm} are capable of producing multiple outputs by sampling from an underlying distribution. Specifically, CVAEs repeatedly sample a latent variable from a prior distribution for decoding. Diffusion models generate multiple outputs by repeatedly sampling independent noise $\mathbf{y}_{(H)}\sim\mathcal{N}(\bm{0},\mathbf{I})$ $M$ times and denoising them resulting in a wide range of varied outputs with complex distributions. With a large number of samples, one can accurately approximate the distribution parameterized by the models. However, in practice, for efficiency, a smaller set of $K$ predictions are used to characterize the performance of the models. Therefore, we require a downsampling (or selection) mechanism to select the most plausible $K$ predictions.

Our goal is to select $K$ most plausible trajectories among $M$ denoised samples given the feature embedding $\bm{C}$, where $K << M$. We achieve this by training scoring network  $g_{\phi}(\cdot)$, parameterized by $\phi$. The scoring network takes $M$ denoised samples and encoder's embedding $\bm{C}$ as input and outputs a score for each of $M$ trajectories conditioned on $\bm{C}$ and the rest of $M-1$ trajectories. For this purpose, we first concatenate each trajectory with feature embedding $\bm{C}$, $\bm{s}_j=\hat{\bm{\tau}}_j\oplus \bm{C}$ for $j=1,\ldots,M$, where $\hat{\bm{\tau}}_j$ is predicted trajectory converted from the $j$th denoised sample $\hat{\mathbf{y}}_{(0)}^j$. We define the matrix $\bm{S} = [\bm{s}_1, ..., \bm{s}_M] \in\mathbb{R}^{M\times (T_f+d_c)}$, which is then fed into a multi-head self-attention block \cite{vaswani2023attention},
$$
\bm{Q}_{i} = \bm{S}\bm{W}^Q_i, \bm{K}_i = \bm{S}\bm{W}^K_i,\bm{V}_i = \bm{S}\bm{W}^V_i \eqno{(6)}
$$\vspace{-0.4cm}
$$
\text{Attn}_{i}(\bm{S}) = \text{softmax}(\frac{1}{\sqrt{d_k}}\bm{Q}_i\bm{K}^T_i)\bm{V}_i \eqno{(7)}
$$\vspace{-0.4cm}
$$
\text{MHA}(\bm{S}) = \text{Concat}(\text{Attn}_1(\bm{S}),\ldots,\text{Attn}_h(\bm{S}))\bm{W}^O \eqno{(8)}
$$where $h$ is the number of attention heads, $\bm{W}^Q_i,\bm{W}^K_i,\bm{W}^V_i\in\mathbb{R}^{(T_f+d_c)\times d_k}$ for $i=1,\ldots,h$, $d_c$ and $d_k$ are the dimensions of the feature embedding and attention, respectively, and $\bm{W}^O\in\mathbb{R}^{h\cdot d_k\times d}$ are the learnable parameters of the multi-head attention module. Next, we apply an MLP layer with residual connection followed by downsampling $d$ to a 1-dimensional score:
$$
\hat{\bm{S}}=g_{\phi}(\bm{S})=(\text{MHA}(\bm{S})+\text{MLP}(\text{MHA}(\bm{S})))\bm{W}^{down} \eqno{(9)}
$$where $\bm{W}^{down}\in\mathbb{R}^{d\times 1}$ is a trainable parameter matrix. Finally, we have predicted relative raw scores $\hat{\bm{S}}\in\mathbb{R}^{^M}$ conditioned on a scene. The scores are normalized using a softmax operation.

\subsection{Training}
As shown in Figure \ref{fig:architecture}, the training process consists of two stages. First, we train denoising module $p_{\theta}$ and encoder $f_{\xi}$, and in the second stage, we train scoring network $g_{\phi}$ with frozen $p_{\theta}$ and $f_{\xi}$.

\textbf{Diffusion}\ \ The model is optimized by maximizing the log-likelihood of the predicted trajectories given the ground truth $\mathbb{E}[\log p_{\theta}(\mathbf{y}_{(0)})]$. Since the exact log-likelihood is intractable, we follow the standard Evidence Lower Bound (ELBO) maximization method and minimize the KL Divergence,
$$
\mathcal{L} = \mathbb{E}_{q,\eta}[D_{KL}(q(\mathbf{y}_{(\eta-1)}|\mathbf{y}_{(\eta)},\mathbf{y}_{(0)})||p_{\theta}(\mathbf{y}_{(\eta-1)}|\mathbf{y}_{(\eta)},\bm{C}))]
$$\vspace{-0.4cm}
$$
=\mathbb{E}_{q,\eta}[D_{KL}(\mathcal{N}(\mathbf{y}_{(\eta-1)};\bm{\mu}_q,\bm{\Sigma}_q(\eta)||\mathcal{N}(\mathbf{y}_{(\eta-1)};\bm{\mu}_{\theta},\bm{\Sigma}_{(\eta)}))]
$$\vspace{-0.4cm}
$$
=\mathbb{E}_{q,\eta}[||\bm{\mu}_{\theta}-\bm{\mu}_q||_2^2] \eqno{(10)}
$$
By utilizing the reparameterization trick, the corresponding optimization problem becomes \cite{ho2020denoisingddpm}:
$$
\mathcal{L}_{MSE}(\theta,\xi) = \mathbb{E}_{\bm{\epsilon}_{(0)},\mathbf{y}_{(0)},\eta}||\bm{\epsilon}_{(0)}-\hat{\bm{\epsilon}}_{(\theta,\xi)}(\mathbf{y}_{(\eta)},\eta,\bm{C})|| \eqno{(11)}
$$where $\bm{\epsilon}_{(0)}\sim\mathcal{N}(\bm{0},\mathbf{I})$, $\mathbf{y}_{(\eta)}=\sqrt{\bar{\alpha_{\eta}}}\mathbf{y}_{(0)}+\sqrt{1-\bar{\alpha}_{\eta}}\bm{\epsilon}_{(0)}$.
In other words, denoising module $\hat{\bm{\epsilon}}_{(\theta,\xi)}$ learns to predict the source noise $\bm{\epsilon}_{(0)}\sim\mathcal{N}(\bm{\epsilon};\bm{0},\mathbf{I})$ that noises $\mathbf{y}_{(0)}$ to $\mathbf{{y}_{(\eta)}}$.

\textbf{Scorer}\ \ We use cross-entropy loss between the predicted scores $\text{softmax}(\hat{\bm{S}})$ and scores calculated using the ground truth future trajectories $\bm{\tau}_{future}$ and the predicted future trajectories $\{\hat{\bm{\tau}}_j\}_{j=1}^M$. A combination of Average Displacement Error (ADE) and final Displacement Error (FDE) is used to calculate the scores,
$$
\psi_j = ADE(\bm{\tau}_{future},\hat{\bm{\tau}}_j) + \lambda FDE(\bm{\tau}_{future},\hat{\bm{\tau}}_j) \eqno{(12)}
$$where $\psi_j$ represents the closeness of $j$th predicted trajectory to the ground truth trajectory, and $\lambda$  balances how metrics ADE and FDE affect the score. $\Psi\in\mathbb{R}^M$ denotes the matrix of $M$ ground truth scores (i.e., $j$th element of $\Psi$ is $\psi_j$). We use a cross-entropy loss to train the scoring network,
$$
\mathcal{L}_{scorer}=\mathcal{L}_{CE}(\text{softmax}(\hat{\bm{S}}),\Psi) \eqno{(13)}
$$

\subsection{Inference}
As shown in Figure \ref{fig:architecture}, inference is a two-stage process: we first oversample $M$ trajectories and denoise them using the denoising module, and then we select the top plausible trajectories.

\textbf{Diffusion}\ \ We begin by sampling $M$ independent Gaussian noises $\{\mathbf{y}_{(H)}^j: \mathbf{y}_{(H)}\sim\mathcal{N}(\bm{0},\mathbf{I})\}_{j=1}^M$. Next, we denoise each $\mathbf{y}_{(H)}^j$ through reverse process $p_{\theta}$. During the reverse process, DDPM \cite{ho2020denoisingddpm} sampling technique repeatedly denoises $\mathbf{y}_{(H)}$ to $\mathbf{y}_{(0)}$ by using equation below for $H$ steps,
$$
\mathbf{y}_{(\eta-1)}=\frac{1}{\sqrt{\alpha_{\eta}}}(\mathbf{y}_{\eta}-\frac{\beta_{\eta}}{\sqrt{1-\bar{\alpha_{\eta}}}}\hat{\bm{\epsilon}}_{(\theta,\xi)}(\mathbf{y}_{(\eta)},\eta,\bm{C}))+\sqrt{\beta_{\eta}}\bm{z} \eqno{(14)}
$$where $\bm{z}\sim\mathcal{N}(\bm{0},\mathbf{I})$.
The downside of the DDPM  sampling is that it requires $H$ denoising steps in order to generate a sample, which is time-consuming and computationally expensive, especially when $H$ is large (usually $H>100$). To mitigate this issue, we use the DDIM \cite{song2022denoisingddim} sampling technique and skip every $\gamma$ step in the reverse process, in which we only iterate $\frac{H}{\gamma}$ steps leading to more efficient and faster operation compared DDPM by a factor of $\gamma$,
$$
\mathbf{y}_{(\eta-1)}=\sqrt{\bar{\alpha}_{(\eta-1)}}(\frac{\mathbf{y}_{(\eta)}-\sqrt{1-\bar{\alpha}_{(\eta)}}\hat{\bm{\epsilon}}_{(\theta,\xi)}(\mathbf{y}_{(\eta)},\eta,\bm{C})}{\sqrt{\bar{\alpha}_{(\eta)}}})
$$\vspace{-0.4cm}
$$
+\sqrt{1-\bar{\alpha}_{(\eta-1)}}\hat{\bm{\epsilon}}_{(\theta,\xi)}(\mathbf{y}_{(\eta)},\eta,\bm{C}) \eqno{(15)}
$$

\textbf{Scoring and selection}\ \ Using the scoring network, we select $K$ out of $M$ oversampled trajectories that are generated by the diffusion model. We first convert predicted displacements, $\hat{\mathbf{y}}_{(0)}=[\hat{\mathbf{y}}_{t+1},\ldots,\hat{\mathbf{y}}_{t+T_f}]$ back to relative positions $\hat{\bm{\tau}}$:
$$
\hat{\bm{\tau}}_{future} = [\hat{\bm{p}}_{t+1},\ldots,\hat{\bm{p}}_{t+T_f}] \eqno{(16)}
$$\vspace{-0.4cm}
$$
\hat{\bm{p}}_{t+j}=\sum_{i'=0}^{j}(\mathbf{R}^{-1})^T\hat{\bm{y}}_{t+i'} \eqno{(17)}
$$where $\mathbf{R}^{-1}$ is the reverse of the rotation matrix used in the preprocessing. Following \cite{zhao2020tnt}, given the scores, we ensure diversity and sufficient multi-modal coverage by applying a non-maximum suppression operation. For this, we sort the trajectories according to their scores, starting with the highest one. If it is greater than the optimized distance threshold $\omega$ from all previously selected trajectories, we select it. We continue until we have $K$ trajectories.


\begin{table*}[ht]
\vspace{+0.2cm}
\caption{Quantitative results on the UCY/ETH benchmark. $\text{minADE}_{20}$/$\text{minFDE}_{20}$ are reported. \textbf{Bold} numbers indicate the best performance for each metric, \underline{underline} indicates the second best performance for each metric. $^*$ indicates that the initial results in the paper are not fairly measured due to the leakage of future paths. We refer to \cite{Liu2023UncertaintyAwarePT} for the corrected results.}\vspace{-0.4cm}
\label{tab:1}
\begin{center}
\resizebox{0.9\textwidth}{!}{%
\begin{tabular}{c||c|c|c|c|c|c}
Method & ETH & Hotel & Univ & Zara1 & Zara2 & Avg\\
\hline
\hline
\rule{0pt}{2ex} 
SocialGAN \cite{gupta2018socialsocialgan} & 0.81/1.52 & 0.72/1.61 & 0.60/1.26 & 0.34/0.69 & 0.42/0.84 & 0.58/1.18\\\hline
SoPhie \cite{sadeghian2018sophie} & 0.70/1.43 & 0.76/1.67 & 0.54/1.24 & 0.30/0.63 & 0.38/0.78 & 0.54/1.15\\\hline
Social-STGCNN  \cite{mohamed2020socialstgcnn} & 0.64/1.11 & 0.49/0.85 & 0.44/0.79 & 0.34/0.53 & 0.30/0.48 & 0.44/0.75\\\hline
Trajectron++$^*$ \cite{salzmann2021trajectron} & 0.67/1.18 & \underline{0.18}/0.28 & \underline{0.30}/0.54 & 0.25/0.41 & \underline{0.18}/\underline{0.32} & 0.32/0.55\\
\hline
\rule{0pt}{2ex}
GroupNet \cite{xu2022groupnet} & \underline{0.46}/\underline{0.73} & \textbf{0.15}/0.25 & \textbf{0.26}/\textbf{0.49} & \textbf{0.21}/\underline{0.39} & \textbf{0.17}/0.33 & \textbf{0.25}/\underline{0.44}\\
\hline
\rule{0pt}{2ex}
MID (Diffusion)$^*$ \cite{gu2022stochasticmid} & 0.54/0.82 & 0.20/0.31 & \underline{0.30}/0.57 & 0.27/0.46 & 0.20/0.37 & 0.30/0.51\\
\hline
\rule{0pt}{2ex}
DICE (ours) & \textbf{0.24}/\textbf{0.34} & \underline{0.18}/\textbf{0.23} & 0.52/0.61 & 0.24/\textbf{0.37} & 0.20/\textbf{0.30} & \underline{0.26}/\textbf{0.35}
\end{tabular}}
\end{center}\vspace{-0.6cm}
\end{table*}

\section{Experiments}

\subsection{Experimental Setup}
\textbf{Datasets.} We evaluate the proposed model on both pedestrian and autonomous driving prediction benchmarks. For the former, we report on \textbf{UCY/ETH} \cite{ucy,4409092eth},  which consists of real pedestrian trajectories at five different locations captured at 2.5 Hz, ETH and HOTEL from ETH dataset, and UNIV, ZARA1, and ZARA2 are from UCY dataset. Following previous works, we use the leave-one-out approach with four sets for training and the remaining set for testing \cite{mao2023leapfrog, xu2022remember}. For autonomous driving, we \textbf{ nuScenes} \cite{nuscenes}, a large-scale real-world autonomous driving dataset, which contains 1000 scenes from Boston and Singapore annotated at 2 Hz. nuScenes provides 2 seconds of history with HD maps and requires 6 seconds of future trajectory to be predicted.
\textbf{Models}. We compare our method against state-of-the-art algorithms on each benchmark. Note that we omit the recent LED \cite{mao2023leapfrog} from the pedestrian benchmark since we were unable to validate the results using the official published code and procedures and no explanations were provided by the authors regarding the discrepancy. We refer to our model as \textbf{DICE} (\textbf{D}iverse dIffusion with s\textbf{C}oring for pr\textbf{E}diction).

\textbf{Metrics} We adopt the standard evaluation metrics including the minimum average/final displacement error over the top $K$ predictions ($\text{minADE}_K$/$\text{minFDE}_K$) on UCY/ETH benchmark. For nuScenes, we also report on miss rate ($\text{MR}_K$) which is the percentage of scenarios where the final points of the $k$ predicted trajectories are more than 2 meters away from the final point of the ground truth trajectory.

\textbf{Implementation Details} 
We train our denoising module for 80 epochs and the scoring network for 20 epochs, using AdamW \cite{loshchilov2019decoupledadamw} optimizer with a learning rate of $5\times10^{-4}$, batch size of 32, and a dropout rate of 0.1. For training the denoising module, we use $H=200$ diffusion steps. When sampling using DDIM \cite{song2022denoisingddim}, we skip over every $\gamma=20$ steps, resulting in only 10 denoising steps to generate each independent trajectory. For training the scoring network, we set the distance metric control weight in $\mathcal{L}_{scorer}$ to be $\lambda=1.5$. Our denoiser consists of 5 transformer layers. All attention modules have $h=4$ heads with 128 hidden dimensions. We set the distance threshold to $\omega=8$ (meters) for the non-maximum suppression operation in inference. All experiments are done on a Tesla V100 GPU. 

\subsection{Comparison to SOTA on Pedestrian Benchmark}
 Table \ref{tab:1} shows the results for $\text{minADE}_{20}$/$\text{minFDE}_{20}$ of models evaluated on UCY/ETH. We can see that the proposed model achieves SOTA performance on 4 out of 5 subsets on $\text{minFDE}$ metrics and overall best performance by a significant margin on \textit{ETH} by improving approx. $50\%$ on both metrics. On average across all subsets, our model is best on $\text{minFDE}$ by improving up to $20\%$ compared to GroupNet while achieving second best on $\text{minADE}$ with a small margin. Such a performance gain is due to our model successfully producing a diverse prediction set that captures a wide range of multi-modal intentions.


\subsection{Qualitative Results}
   \begin{figure}[h]
      \begin{center}
      \includegraphics[scale=0.2]{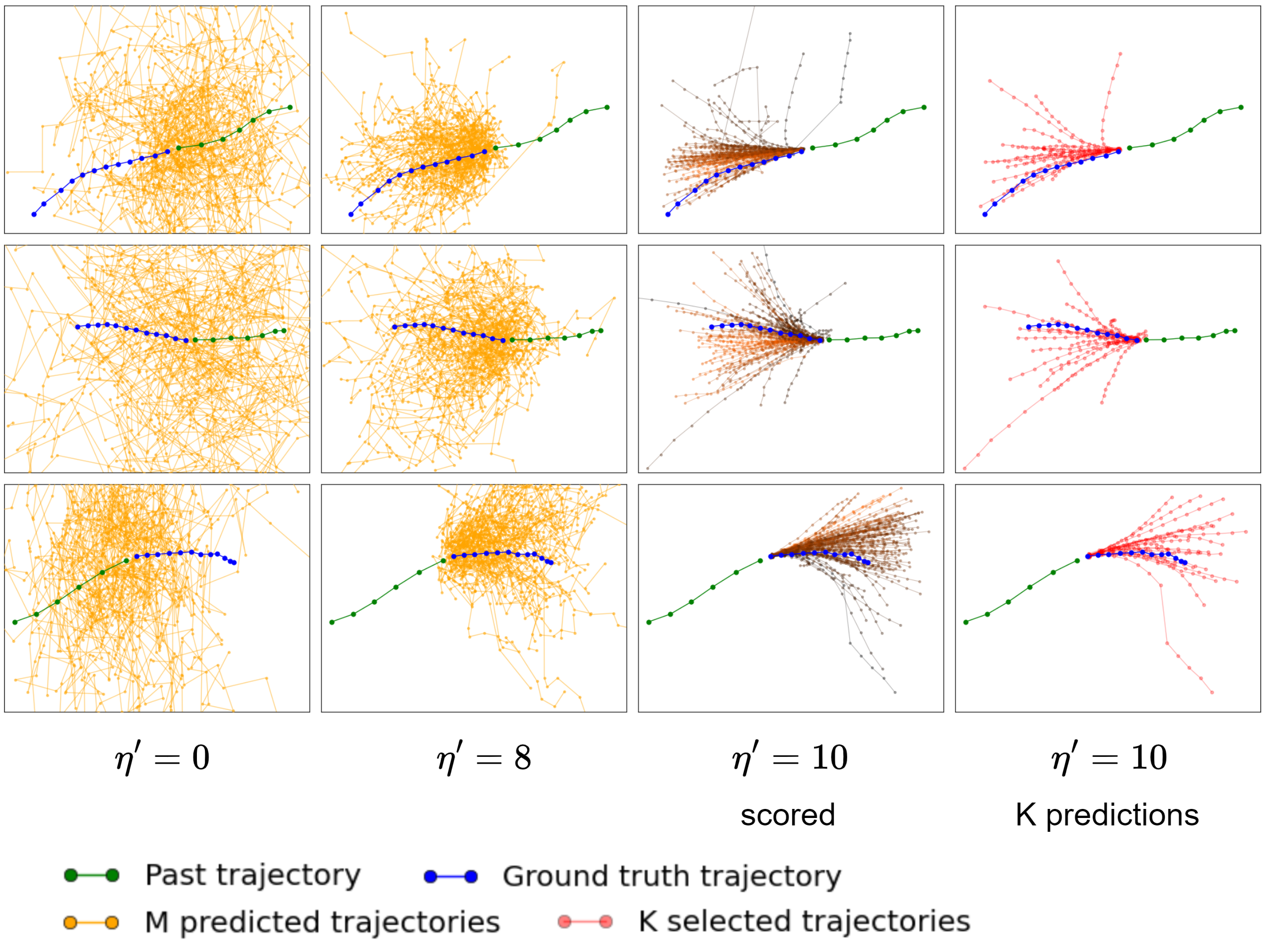}\vspace{-0.3cm}
      \caption{Visualization of generated trajectories from DDIM sampling at each diffusion step $\eta'$. The plots in the last two columns show the generated trajectories after scoring and after applying the selection algorithm respectively. Lighter colours reflect higher scores.}\vspace{-0.6cm}
      \label{fig:qual}
      \end{center}
   \end{figure}
   
Figure \ref{fig:qual} illustrates the reverse process of our model on a subset of scenes from the UCY/ETH dataset. $\eta'=\frac{\eta}{\gamma}$ denotes the diffusion steps taken by DDIM sampling, where DDPM sampling steps being $H=200$ and the number of steps skipped being $\gamma=20$, which results in the total diffusion steps taken by DDIM sampling being $10$. We plot $M=100$ generated trajectories after being scored, and $K=20$ final selected. The last column in Figure \ref{fig:qual} shows that the selected trajectories using the scoring network are very diverse, but also more dense around the ground truth trajectory.
\subsection{Comparison to SOTA on Autonomous Driving Benchmark}
\begin{table*}[h]
\vspace{+0.2cm}
\caption{Quantitative results on the nuScenes benchmark. \textbf{Bold} numbers indicate the best performance for each metric, \underline{underline} indicates the second best performance for each metric.} \vspace{-0.4cm}
\label{tab:nuscenes}
\begin{center}
\resizebox{0.9\textwidth}{!}{%
\begin{tabular}{c||c|c|c|c|c|c|c}
Method & $\text{minADE}_5$ & $\text{minADE}_{10}$ & $\text{minFDE}_1$ & $\text{minFDE}_5$ & $\text{minFDE}_{10}$ & $\text{MR}_5$ & $\text{MR}_{10}$\\
\hline
\hline
\rule{0pt}{2ex}
CoverNet \cite{phanminh2020covernet}  & 1.96 & 1.48  & 9.26 & - & - & 0.67  & - \\
\hline
\rule{0pt}{2ex}
Trajectron++ \cite{salzmann2021trajectron} & 1.88  & 1.51  & 9.52 & - & - & 0.70 & 0.57 \\
\hline
\rule{0pt}{2ex}
AgentFormer \cite{yuan2021agentformer} & 1.86  & 1.45  & - & 3.89 & 2.86 & - & - \\
\hline
\rule{0pt}{2ex}
SG-Net \cite{zhang2019sgnet} & 1.86  & 1.40  & 9.25 & - & - & 0.67 & 0.52 \\
\hline
\rule{0pt}{2ex}
MHA-JAM \cite{messaoud2020trajectorymhajam} & 1.81 & 1.24 & 8.57 & \underline{3.72} & \textbf{2.21} & 0.59 & 0.46 \\
\hline
\rule{0pt}{2ex}
CXX \cite{luo2020probabilisticcxx}& 1.63  & 1.29 & - & - & - & 0.69 & 0.60 \\
\hline
\rule{0pt}{2ex}
P2T \cite{deo2021trajectoryp2t} & 1.45  & 1.16 & 10.45 & - & - & 0.64 & 0.46 \\
\hline
\rule{0pt}{2ex}
GOHOME \cite{gilles2021gohome} & \underline{1.42}  & \underline{1.15} & \textbf{6.99} & - & - &\underline{0.57} & 0.47 \\
\hline
\rule{0pt}{2ex}
PGP \cite{deo2021multimodalpgp} &\textbf{1.30}  & \textbf{1.00} & \underline{7.17} & - & - & 0.61 & \underline{0.37} \\
\hline
\rule{0pt}{2ex}
DICE (ours) (w/o scoring) & 2.03 & 1.68 & 8.90 & 3.83 & 2.80 & 0.55 & 0.40 \\
\hline
\rule{0pt}{2ex}
DICE (ours) (w/ scoring) & 1.76 & 1.44 & 7.74 & \textbf{3.70} & \underline{2.67} & \textbf{0.53} & \textbf{0.34}
\end{tabular}}
\end{center}\vspace{-0.4cm}
\end{table*}

To further highlight the effectiveness of the proposed model, we evaluate our approach on the nuScenes autonomous driving benchmark. We report the results for $\text{minADE}_{K}$, $\text{minFDE}_{K}$, $\text{MR}_{K}$, for $K=1,5,10$. The results are summarized in Table \ref{tab:nuscenes}. Here, we can see that our method is particularly effective in predicting endpoints (targets) as it achieves an improvement of up to $9\%$ on miss rate while ranking first and second on $\text{minFDE}_{5}$ and $\text{minFDE}_{10}$, respectively. Again, this confirms the ability of our prediction set to capture intention points. This should be noted that in the autonomous driving domain, more emphasis is often given to final error, variations of which are used for ranking of prediction models in more recent benchmarks, such as \cite{chang2019argoverse,sun2020scalability}. In terms of $\text{minADE}$ our model lags behind, which can be due to the tendency of the model to generate more low curvature trajectories, hence, causing higher average error in driving scenes as they may contain turns.

\subsection{Ablation Studies}
\begin{table}[h]
\caption{Ablation study on post-processing of trajectory selection method. Results are reported using $\text{minADE}_{20}$/$\text{minFDE}_{20}$ on UCY/ETH.}\vspace{-0.4cm}
\label{tab:pptsm}
\begin{center}
\resizebox{0.9\columnwidth}{!}{%
\begin{tabular}{c||c|c|c}
\  & Random & Clustering & Scoring+NMS\\
\hline
\hline
\rule{0pt}{2ex}
ETH  & 0.27/0.42 & \textbf{0.24}/0.35 & \textbf{0.24}/\textbf{0.34} \\
\hline
\rule{0pt}{2ex}
Hotel & 0.20/0.28 & 0.19/0.25  & \textbf{0.18}/\textbf{0.23} \\
\hline
\rule{0pt}{2ex}
Univ & 0.46/\textbf{0.52} & \textbf{0.45}/\textbf{0.52} & 0.52/0.61 \\
\hline
\rule{0pt}{2ex}
Zara1 & 0.26/0.42  & \textbf{0.24}/\textbf{0.37} & \textbf{0.24}/\textbf{0.37} \\
\hline
\rule{0pt}{2ex}
Zara2 & 0.23/0.40 & 0.22/0.36  & \textbf{0.20}/\textbf{0.30} \\
\hline
\rule{0pt}{2ex}
AVG  & 0.28/0.41 & 0.27/0.37 & \textbf{0.26}/\textbf{0.35}
\end{tabular}}
\end{center}\vspace{-0.5cm}
\end{table}

\textbf{Effect of the scoring network.}\ \ We evaluate the effect of oversampling and then undersampling trajectories with our scoring model and non-maximum suppression by comparing against 2 baselines. These encompass randomly generating $K$ trajectories directly from our denoiser, and an intelligent post-processing selection algorithm implemented in \cite{jiang2023motiondiffuser} and partially inspired from \cite{varadarajan2021multipath}. Let $\hat{T}_{1:K}=\{\hat{\bm{\tau}}_{future}^{j}\}_{j=1}^K$ be the selected $K$ predicted trajectories, and $\hat{T}_{1:M}=\{\hat{\bm{\tau}}_{future}^{j}\}_{j=1}^M$ be the oversampled set of $M$ trajectories. We attempt to select $\hat{T}_{1:K}$ so that the maximum number of elements in $\hat{T}_{1:M}$ lie within distance $r$ of any of the $\hat{T}_{1:K}$ elements.
$$
\hat{T}_{1:K}= \argmax_{\hat{T}_{1:K}} \sum_{j=1}^{M}  \;  \maxover_{\hat{T}_{i} \in \hat{T}_{1:K} }  \: 
 \mathbb{I} \, ( \, dist(\hat{T}_{i},\hat{T}_{j})<r)
$$where $dist(.)$ is the distance function for which we use ADE, and $r$ is an adjustable threshold. We approximate our solution in a greedy fashion, iteratively adding one trajectory to $\hat{T}_{1:K}$ until we have $K$ trajectories. The trajectory we select at each step is the one that maximizes the above argmax objective when added to the current set. 

As shown in Table \ref{tab:pptsm}, selecting trajectories using the scoring network with non-maximum suppression achieves the best results, with on average $7\%/15\%$ improvement on $\text{minADE}_{20}$/$\text{minFDE}_{20}$ compared to the case where we randomly sample $20$ trajectories directly from the denoiser. It also performs up to $4\%/5\%$ better than our post-processing selection baseline. The scoring network, however, negatively affects the performance on the \textit{Univ} subset. We hypothesize that the substantially smaller size of the train set compared to the test set for  \textit{Univ}  may have lead to underfitting.


 \begin{figure}[thpb]
 \centering
       \includegraphics[width=1\columnwidth]{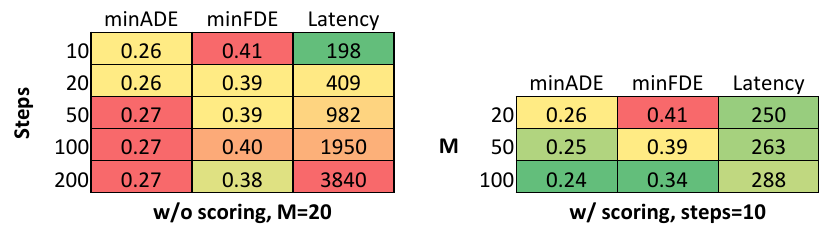}\vspace{-0.2cm}
       \caption{Ablation study on number of denoising \textit{Steps} and oversampling size (\textit{M}). Metrics are $\text{minADE}_{20}$/$\text{minFDE}_{20}$/latency(ms) and reported on ETH. Colors indicate better (green) and worse (red) values and are assigned columnwise for each metric and normalized across both tables.}
      \label{fig:abl_scale}\vspace{-0.3cm}
    \end{figure}

\textbf{Steps vs sampling}\ \
We study the effect of the number of steps and samples on accuracy and latency. The results are illustrated in Figure \ref{fig:abl_scale}. On the left, as one would expect, increasing the number of denoising steps improves the performance, although only minFDE metric. This, however, comes at a significant cost of increasing the latency by as much as 19 times. On the other hand, as shown in the table on the right, we can see that much better improvement can be achieved by simply increasing the number of samples. Increasing the number of samples by five-fold, only increases latency by $15\%$ while improving the performance by up to $21\%$. This highlights the effectiveness of oversampling and the proposed scoring method to improve accuracy.


\section{Conclusion}
We presented a novel model for road user trajectory prediction, leveraging the capabilities of diffusion models. Our approach benefited from an efficient sampling approach resulting in significant speed-up. This allows our model to oversample trajectory distributions to better capture the space of possibilities. Furthermore, we proposed a scoring network that ranks the sampled trajectories in order to select the most plausible ones. We conducted extensive evaluations on the pedestrian and autonomous driving benchmark datasets and showed that our model achieves state-of-the-art performance on a number of subsets and metrics, in particular $\text{minFDE}$ and $\text{MR}$. In addition, we conducted ablative studies to highlight the effectiveness of the proposed scoring scheme and oversampling on boosting the accuracy of predictions. 

{\small
	\bibliographystyle{IEEEtranS}
	\bibliography{ref}

\begin{thebibliography}{10}
\providecommand{\url}[1]{#1}
\csname url@samestyle\endcsname
\providecommand{\newblock}{\relax}
\providecommand{\bibinfo}[2]{#2}
\providecommand{\BIBentrySTDinterwordspacing}{\spaceskip=0pt\relax}
\providecommand{\BIBentryALTinterwordstretchfactor}{4}
\providecommand{\BIBentryALTinterwordspacing}{\spaceskip=\fontdimen2\font plus
\BIBentryALTinterwordstretchfactor\fontdimen3\font minus
  \fontdimen4\font\relax}
\providecommand{\BIBforeignlanguage}[2]{{%
\expandafter\ifx\csname l@#1\endcsname\relax
\typeout{** WARNING: IEEEtranS.bst: No hyphenation pattern has been}%
\typeout{** loaded for the language `#1'. Using the pattern for}%
\typeout{** the default language instead.}%
\else
\language=\csname l@#1\endcsname
\fi
#2}}
\providecommand{\BIBdecl}{\relax}
\BIBdecl

\bibitem{alahi2016social}
A.~Alahi, K.~Goel, V.~Ramanathan, A.~Robicquet, L.~Fei-Fei, and S.~Savarese,
  ``Social {LSTM}: Human trajectory prediction in crowded spaces,'' in
  \emph{CVPR}, 2016.

\bibitem{amirloo2022latentformer}
E.~Amirloo, A.~Rasouli, P.~Lakner, M.~Rohani, and J.~Luo, ``{LatentFormer}:
  Multi-agent transformer-based interaction modeling and trajectory
  prediction,'' \emph{arXiv:2203.01880}, 2022.

\bibitem{nuscenes}
H.~Caesar, V.~Bankiti, A.~H. Lang, S.~Vora, V.~E. Liong, Q.~Xu, A.~Krishnan,
  Y.~Pan, G.~Baldan, and O.~Beijbom, ``{nuScenes}: A multimodal dataset for
  autonomous driving,'' in \emph{CVPR}, 2020.

\bibitem{chang2019argoverse}
M.-F. Chang, J.~Lambert, P.~Sangkloy, J.~Singh, S.~Bak, A.~Hartnett, D.~Wang,
  P.~Carr, S.~Lucey, D.~Ramanan \emph{et~al.}, ``Argoverse: {3D} tracking and
  forecasting with rich maps,'' in \emph{CVPR}, 2019.

\bibitem{chen2022personalized}
G.~Chen, J.~Li, N.~Zhou, L.~Ren, and J.~Lu, ``Personalized trajectory
  prediction via distribution discrimination,'' in \emph{ICCV}, 2021.

\bibitem{choi2020drogon}
C.~Choi, S.~Malla, A.~Patil, and J.~H. Choi, ``{DROGON}: A trajectory
  prediction model based on intention-conditioned behavior reasoning,'' in
  \emph{CoRL}, 2021.

\bibitem{dendorfer2021mggan}
P.~Dendorfer, S.~Elflein, and L.~Leal-Taixé, ``{MG-GAN}: A multi-generator
  model preventing out-of-distribution samples in pedestrian trajectory
  prediction,'' in \emph{ICCV}, 2021.

\bibitem{deo2021trajectoryp2t}
N.~Deo and M.~M. Trivedi, ``Trajectory forecasts in unknown environments
  conditioned on grid-based plans,'' \emph{arXiv:2001.00735}, 2020.

\bibitem{deo2021multimodalpgp}
N.~Deo, E.~M. Wolff, and O.~Beijbom, ``Multimodal trajectory prediction
  conditioned on lane-graph traversals,'' in \emph{CoRL}, 2022.

\bibitem{4409092eth}
A.~Ess, B.~Leibe, and L.~Van~Gool, ``Depth and appearance for mobile scene
  analysis,'' in \emph{ICCV}, 2007.

\bibitem{Fang_2020_CVPRtpnet}
L.~Fang, Q.~Jiang, J.~Shi, and B.~Zhou, ``Tpnet: Trajectory proposal network
  for motion prediction,'' in \emph{CVPR}, 2020.

\bibitem{fernando2017soft}
T.~Fernando, S.~Denman, S.~Sridharan, and C.~Fookes, ``Soft+ hardwired
  attention: An lstm framework for human trajectory prediction and abnormal
  event detection,'' \emph{Neural Networks}, vol. 108, pp. 466--478, 2018.

\bibitem{gao2020vectornet}
J.~Gao, C.~Sun, H.~Zhao, Y.~Shen, D.~Anguelov, C.~Li, and C.~Schmid,
  ``{VectorNet}: Encoding hd maps and agent dynamics from vectorized
  representation,'' in \emph{CVPR}, 2020.

\bibitem{gilles2021gohome}
T.~Gilles, S.~Sabatini, D.~Tsishkou, B.~Stanciulescu, and F.~Moutarde,
  ``{GOHOME}: Graph-oriented heatmap output for future motion estimation,'' in
  \emph{ICRA}, 2022.

\bibitem{gu2021densetnt}
J.~Gu, C.~Sun, and H.~Zhao, ``{DenseTNT}: End-to-end trajectory prediction from
  dense goal sets,'' in \emph{ICCV}, 2021.

\bibitem{gu2022stochasticmid}
T.~Gu, G.~Chen, J.~Li, C.~Lin, Y.~Rao, J.~Zhou, and J.~Lu, ``Stochastic
  trajectory prediction via motion indeterminacy diffusion,'' in \emph{CVPR},
  2022.

\bibitem{gupta2018socialsocialgan}
A.~Gupta, J.~Johnson, L.~Fei-Fei, S.~Savarese, and A.~Alahi, ``{Social GAN}:
  Socially acceptable trajectories with generative adversarial networks,'' in
  \emph{CVPR}, 2018.

\bibitem{ho2022imagen}
J.~Ho, W.~Chan, C.~Saharia, J.~Whang, R.~Gao, A.~Gritsenko, D.~P. Kingma,
  B.~Poole, M.~Norouzi, D.~J. Fleet \emph{et~al.}, ``Imagen video: High
  definition video generation with diffusion models,'' \emph{arXiv:2210.02303},
  2022.

\bibitem{ho2020denoisingddpm}
J.~Ho, A.~Jain, and P.~Abbeel, ``Denoising diffusion probabilistic models,'' in
  \emph{NeurIPS}, 2020.

\bibitem{Huang_2019_ICCV}
Y.~Huang, H.~Bi, Z.~Li, T.~Mao, and Z.~Wang, ``{STGAT}: Modeling
  spatial-temporal interactions for human trajectory prediction,'' in
  \emph{ICCV}, 2019.

\bibitem{ivanovic2019trajectron}
B.~Ivanovic and M.~Pavone, ``The {Trajectron}: Probabilistic multi-agent
  trajectory modeling with dynamic spatiotemporal graphs,'' in \emph{ICCV},
  2019.

\bibitem{jiang2023motiondiffuser}
C.~M. Jiang, A.~Cornman, C.~Park, B.~Sapp, Y.~Zhou, and D.~Anguelov,
  ``{MotionDiffuser}: Controllable multi-agent motion prediction using
  diffusion,'' in \emph{CVPR}, 2023.

\bibitem{kong2021diffwave}
Z.~Kong, W.~Ping, J.~Huang, K.~Zhao, and B.~Catanzaro, ``Diffwave: A versatile
  diffusion model for audio synthesis,'' \emph{arXiv:2009.09761}, 2020.

\bibitem{kosaraju2019socialbigat}
V.~Kosaraju, A.~Sadeghian, R.~Martín-Martín, I.~Reid, S.~H. Rezatofighi, and
  S.~Savarese, ``{Social-BiGAT}: Multimodal trajectory forecasting using
  bicycle-gan and graph attention networks,'' in \emph{NeurIPS}, 2019.

\bibitem{lee2017desire}
N.~Lee, W.~Choi, P.~Vernaza, C.~B. Choy, P.~H.~S. Torr, and M.~Chandraker,
  ``{DESIRE}: Distant future prediction in dynamic scenes with interacting
  agents,'' in \emph{CVPR}, 2017.

\bibitem{ucy}
A.~Lerner, Y.~Chrysanthou, and D.~Lischinski, ``Crowds by example,''
  \emph{Computer Graphics Forum}, vol.~26, no.~3, pp. 655--664, 2007.

\bibitem{Liu2023UncertaintyAwarePT}
Y.~Liu, Z.~Ye, B.~Li, and L.~Yao, ``Uncertainty-aware pedestrian trajectory
  prediction via distributional diffusion,'' \emph{ArXiv}, 2023.

\bibitem{Liu_2021_ICCV}
Y.~Liu, Q.~Yan, and A.~Alahi, ``{Social NCE}: Contrastive learning of
  socially-aware motion representations,'' in \emph{ICCV}, 2021.

\bibitem{loshchilov2019decoupledadamw}
I.~Loshchilov and F.~Hutter, ``Decoupled weight decay regularization,''
  \emph{arXiv:1711.05101}, 2017.

\bibitem{luo2020probabilisticcxx}
C.~Luo, L.~Sun, D.~Dabiri, and A.~Yuille, ``Probabilistic multi-modal
  trajectory prediction with lane attention for autonomous vehicles,'' in
  \emph{IROS}, 2020.

\bibitem{luo2021diffusion3dcloud}
S.~Luo and W.~Hu, ``Diffusion probabilistic models for 3d point cloud
  generation,'' in \emph{CVPR}, 2021.

\bibitem{mao2023leapfrog}
W.~Mao, C.~Xu, Q.~Zhu, S.~Chen, and Y.~Wang, ``Leapfrog diffusion model for
  stochastic trajectory prediction,'' in \emph{CVPR}, 2023.

\bibitem{messaoud2020trajectorymhajam}
K.~Messaoud, N.~Deo, M.~M. Trivedi, and F.~Nashashibi, ``Trajectory prediction
  for autonomous driving based on multi-head attention with joint agent-map
  representation,'' in \emph{IV}, 2021.

\bibitem{mohamed2020socialstgcnn}
A.~Mohamed, K.~Qian, M.~Elhoseiny, and C.~Claudel, ``{Social-STGCNN}: A social
  spatio-temporal graph convolutional neural network for human trajectory
  prediction,'' in \emph{CVPR}, 2020.

\bibitem{nayakanti2023wayformer}
N.~Nayakanti, R.~Al-Rfou, A.~Zhou, K.~Goel, K.~S. Refaat, and B.~Sapp,
  ``Wayformer: Motion forecasting via simple \& efficient attention networks,''
  in \emph{ICRA}, 2023.

\bibitem{park2020diverse}
S.~H. Park, G.~Lee, J.~Seo, M.~Bhat, M.~Kang, J.~Francis, A.~Jadhav, P.~P.
  Liang, and L.-P. Morency, ``Diverse and admissible trajectory forecasting
  through multimodal context understanding,'' in \emph{ECCV}, 2020.

\bibitem{phanminh2020covernet}
T.~Phan-Minh, E.~C. Grigore, F.~A. Boulton, O.~Beijbom, and E.~M. Wolff,
  ``{CoverNet}: Multimodal behavior prediction using trajectory sets,'' in
  \emph{CVPR}, 2020.

\bibitem{rasouli2023pedformer}
A.~Rasouli and I.~Kotseruba, ``Pedformer: Pedestrian behavior prediction via
  cross-modal attention modulation and gated multitask learning,'' in
  \emph{ICRA}, 2023.

\bibitem{Rasouli_2021_ICCV}
A.~Rasouli, M.~Rohani, and J.~Luo, ``Bifold and semantic reasoning for
  pedestrian behavior prediction,'' in \emph{ICCV}, 2021.

\bibitem{rempe2023trace}
D.~Rempe, Z.~Luo, X.~B. Peng, Y.~Yuan, K.~Kitani, K.~Kreis, S.~Fidler, and
  O.~Litany, ``Trace and pace: Controllable pedestrian animation via guided
  trajectory diffusion,'' in \emph{CVPR}, 2023.

\bibitem{rombach2022high}
R.~Rombach, A.~Blattmann, D.~Lorenz, P.~Esser, and B.~Ommer, ``High-resolution
  image synthesis with latent diffusion models,'' in \emph{CVPR}, 2022.

\bibitem{sadeghian2018sophie}
A.~Sadeghian, V.~Kosaraju, A.~Sadeghian, N.~Hirose, S.~H. Rezatofighi, and
  S.~Savarese, ``{SoPhie}: An attentive gan for predicting paths compliant to
  social and physical constraints,'' in \emph{CVPR}, 2019.

\bibitem{saharia2022photorealistic}
C.~Saharia, W.~Chan, S.~Saxena, L.~Li, J.~Whang, E.~L. Denton, K.~Ghasemipour,
  R.~Gontijo~Lopes, B.~Karagol~Ayan, T.~Salimans \emph{et~al.},
  ``Photorealistic text-to-image diffusion models with deep language
  understanding,'' in \emph{NeurIPS}, 2022.

\bibitem{salzmann2021trajectron}
T.~Salzmann, B.~Ivanovic, P.~Chakravarty, and M.~Pavone, ``Trajectron++:
  Dynamically-feasible trajectory forecasting with heterogeneous data,'' in
  \emph{ECCV}, 2020.

\bibitem{Sohn2015LearningSOcvae}
K.~Sohn, H.~Lee, and X.~Yan, ``Learning structured output representation using
  deep conditional generative models,'' in \emph{NIPS}, 2015.

\bibitem{song2022denoisingddim}
J.~Song, C.~Meng, and S.~Ermon, ``Denoising diffusion implicit models,''
  \emph{arXiv:2010.02502}, 2020.

\bibitem{sun2020reciprocal}
H.~Sun, Z.~Zhao, and Z.~He, ``Reciprocal learning networks for human trajectory
  prediction,'' in \emph{CVPR}, 2020.

\bibitem{sun2020recursive}
J.~Sun, Q.~Jiang, and C.~Lu, ``Recursive social behavior graph for trajectory
  prediction,'' in \emph{CVPR}, 2020.

\bibitem{sun2020scalability}
P.~Sun, H.~Kretzschmar, X.~Dotiwalla, A.~Chouard, V.~Patnaik, P.~Tsui, J.~Guo,
  Y.~Zhou, Y.~Chai, B.~Caine \emph{et~al.}, ``Scalability in perception for
  autonomous driving: Waymo open dataset,'' in \emph{CVPR}, 2020.

\bibitem{thanhtung2020catastrophic}
H.~Thanh-Tung and T.~Tran, ``On catastrophic forgetting in generative
  adversarial networks,'' \emph{arXiv:1807.04015}, 2018.

\bibitem{varadarajan2021multipath}
B.~Varadarajan, A.~Hefny, A.~Srivastava, K.~S. Refaat, N.~Nayakanti,
  A.~Cornman, K.~Chen, B.~Douillard, C.~P. Lam, D.~Anguelov, and B.~Sapp,
  ``{MultiPath++}: Efficient information fusion and trajectory aggregation for
  behavior prediction,'' in \emph{ICRA}, 2022.

\bibitem{vaswani2023attention}
A.~Vaswani, N.~Shazeer, N.~Parmar, J.~Uszkoreit, L.~Jones, A.~N. Gomez,
  L.~Kaiser, and I.~Polosukhin, ``Attention is all you need,'' in
  \emph{NeurIPS}, 2017.

\bibitem{vemula2018social}
A.~Vemula, K.~Muelling, and J.~Oh, ``Social attention: Modeling attention in
  human crowds,'' in \emph{ICRA}, 2018.

\bibitem{xu2022groupnet}
C.~Xu, M.~Li, Z.~Ni, Y.~Zhang, and S.~Chen, ``{GroupNet}: Multiscale hypergraph
  neural networks for trajectory prediction with relational reasoning,'' in
  \emph{CVPR}, 2022.

\bibitem{xu2022remember}
C.~Xu, W.~Mao, W.~Zhang, and S.~Chen, ``Remember intentions:
  Retrospective-memory-based trajectory prediction,'' in \emph{CVPR}, 2022.

\bibitem{yang2022diffusion}
R.~Yang, P.~Srivastava, and S.~Mandt, ``Diffusion probabilistic modeling for
  video generation,'' \emph{arXiv:2203.09481}, 2022.

\bibitem{ye2021tpcn}
M.~Ye, T.~Cao, and Q.~Chen, ``{TPCN}: Temporal point cloud networks for motion
  forecasting,'' in \emph{CVPR}, 2021.

\bibitem{yu2020spatiotemporal}
C.~Yu, X.~Ma, J.~Ren, H.~Zhao, and S.~Yi, ``Spatio-temporal graph transformer
  networks for pedestrian trajectory prediction,'' in \emph{ECCV}, 2020.

\bibitem{yuan2021agentformer}
Y.~Yuan, X.~Weng, Y.~Ou, and K.~Kitani, ``{AgentFormer}: Agent-aware
  transformers for socio-temporal multi-agent forecasting,'' in \emph{ICCV},
  2021.

\bibitem{zhang2019srlstm}
P.~Zhang, W.~Ouyang, P.~Zhang, J.~Xue, and N.~Zheng, ``{SR-LSTM}: State
  refinement for lstm towards pedestrian trajectory prediction,'' in
  \emph{CVPR}, 2019.

\bibitem{zhang2019sgnet}
Z.~Zhang, Y.~Wu, J.~Zhou, S.~Duan, H.~Zhao, and R.~Wang, ``{SG-Net}:
  Syntax-guided machine reading comprehension,'' in \emph{AAAI}, 2020.

\bibitem{zhao2020tnt}
H.~Zhao, J.~Gao, T.~Lan, C.~Sun, B.~Sapp, B.~Varadarajan, Y.~Shen, Y.~Shen,
  Y.~Chai, C.~Schmid, C.~Li, and D.~Anguelov, ``{TNT}: Target-driven trajectory
  prediction,'' in \emph{CoRL}, 2021.

\bibitem{zhao2017deeper}
S.~Zhao, J.~Song, and S.~Ermon, ``Towards deeper understanding of variational
  autoencoding models,'' \emph{arXiv:1702.08658}, 2017.

\bibitem{hivt}
Z.~Zhou, L.~Ye, J.~Wang, K.~Wu, and K.~Lu, ``{HiVT}: Hierarchical vector
  transformer for multi-agent motion prediction,'' in \emph{CVPR}, 2022.

\end{thebibliography}
}

\end{document}